\documentclass[sigconf,nonacm]{acmart}

\usepackage[british]{babel}
\usepackage{booktabs}
\usepackage{multirow}
\usepackage{amsmath}
\usepackage{microtype}
\usepackage{balance}

\AtBeginDocument{%
  }

\begin{document}

\title[HiPAD: Unified Multimodal Depression Detection, a Personality Ablation, and LLM Calibration Limits]{HiPAD: A Unified Gait, Audio-Visual, and Personality Fusion Network for Depression Detection, with a Hierarchical Personality Ablation and the Calibration Limits of Domain-Adapted LLMs}

\author{Karol Chlasta}
\orcid{0000-0002-6539-566X}
\affiliation{%
  \institution{Kozminski University}
  \city{Warsaw}
  \country{Poland}}
\affiliation{%
  \institution{WarsawIQ}
  \city{Warsaw}
  \country{Poland}}
\email{kchlasta@kozminski.edu.pl}
\email{karol.chlasta@warsawiq.com}

\author{Piotr Struzik}
\orcid{0009-0000-0266-9144}
\affiliation{%
  \institution{WarsawIQ}
  \city{Warsaw}
  \country{Poland}}
\email{piotr.struzik@warsawiq.com}

\begin{abstract}
We present \textbf{HiPAD}, a single fusion network submitted to all
six sub-tracks of the MPDD-AVG 2026 depression-detection challenge,
covering the Elder and Young cohorts and three modality sets---gait,
audio-visual, and both---each combined with personality descriptions.
The network encodes each available modality with a shared BiLSTM,
zero-pads absent modalities, and applies cross-modal attention only
when at least two of gait, audio, and video are present. Personality,
which dominates our results, is kept out of the attention and fused
by concatenation.
Because the architecture is identical in every sub-track, differences
between sub-tracks reflect the data rather than the model, and we use
this to study three questions.
Our main finding is that the personality description---roughly 150
tokens of text about who the subject is---carries more depression
signal than audio, video, and gait combined: gait plus personality
alone (TLS 0.647) beats every configuration that includes audio and
video.
First, a four-level ablation of the personality description (L0--L3)
quantifies this: removing every personality-derived feature,
including the Big Five scores that earlier MPDD ablations keep,
lowers the Track-Level Score by 0.499 to 0.507 and leaves the model
at chance.
Second, domain-adapting the personality encoder with LoRA (7B) or QLoRA
(72B) on PHQ-8/9 data raises binary Cohen's~$\kappa$ by 0.13 but lowers
the overall score by distorting the PHQ-9 regression; a frozen
RoBERTa-large encoder does better at this sample size.
Third, at fixed architecture, handcrafted clinical features (eGeMAPS,
OpenFace action units) outperform generic deep features (wav2vec2,
DenseNet) on the audio-visual tracks.
The unified system reaches a cross-track G-Score of 0.482; with fewer
than 100 subjects per cohort, the limiting factors are the size of the
representation and the calibration of the outputs, not the choice of
architecture.
\end{abstract}



\maketitle

\begingroup\small\itshape\noindent
\rule{\columnwidth}{0.4pt}\\[2pt]
This is the authors' version of the work, posted here for personal use and
extended with material omitted from the conference version for length.
The definitive Version of Record was published in \emph{Proceedings of the
34th ACM International Conference on Multimedia (MM~'26)},
\url{https://doi.org/10.1145/3767308.3836618}.\\[2pt]
\rule{\columnwidth}{0.4pt}
\endgroup

\section{Introduction}

Depression affects over 280~million people globally~\cite{who2023}.
Automated assessment benefits from multiple behavioural modalities:
personality traits capture stable dispositional risk
factors~\cite{kotov2010}, while audio-visual cues and gait patterns
encode momentary psychomotor symptoms.
The MPDD-AVG 2026 challenge~\cite{fu2026mpddavg,mpdd2026} organises six
sub-tracks across Elder ($\geq$60 years) and Young cohorts, combining
gait (IMU), audio-visual interviews, and---uniquely---free-text
\emph{personality descriptions} encoding Big Five traits, demographics,
and disease history.

Challenge entries are typically specialised per sub-track, which
precludes cross-track conclusions; and prior personality-aware MPDD
systems (P3HF~\cite{p3hf}, Fosafer~\cite{fosafer}) measure the
personality modality against \emph{soft}-ablation baselines that
retain numerical Big Five scores.
We address both gaps with one system evaluated on all six sub-tracks,
and make four contributions:

\begin{enumerate}
\item \textbf{A unified variable-modality fusion network.}
  HiPAD encodes whichever modalities a sub-track provides with a shared
  BiLSTM design, activates cross-modal attention only between $\geq$2
  non-personality modalities, and fuses the dominant personality
  stream by concatenation only (Section~\ref{sec:arch}); one
  capacity-matched configuration ($h{=}128$) covers five of six
  sub-tracks unchanged, reaching a cross-track G-Score of 0.482.

\item \textbf{A true hierarchical personality ablation.}
  A four-level decomposition (L0--L3, Section~\ref{sec:hierarchy})
  whose L0 baseline removes \emph{all} personality-derived features
  shows the personality-description modality contributes +0.499--0.507
  average Track-Level Score, and that L0 models are at chance
  (Section~\ref{sec:hier-results}).

\item \textbf{The calibration limit of domain-adapted LLMs.}
  LoRA 7B and QLoRA 72B personality encoders raise binary $\kappa$ by
  $+0.13$ but lower the overall score through PHQ-9 regression
  calibration mismatch, isolated with hybrid task-head experiments
  (Section~\ref{sec:calib}).

\item \textbf{Cohort-dependent feature selection under one
  architecture.}
  Clinically-motivated features (eGeMAPS~\cite{eyben2016},
  OpenFace~\cite{baltrusaitis2018} action units) beat generic deep
  features in every same-architecture audio-visual comparison
  (Section~\ref{sec:audio}), and one capacity-vs-$n$ analysis explains
  why larger hidden widths, larger embeddings, and extra modalities
  all fail to help at $n{=}87$--$88$ (Section~\ref{sec:capacity}).
\end{enumerate}

\section{Related Work}
\label{sec:rw}

Depression detection has been studied using audio
features~\cite{eyben2016,chlasta2019speech,chlasta2021spontaneous},
video features~\cite{baltrusaitis2018}, text
transcripts~\cite{depresym}, audio-visual shared
tasks~\cite{avec2016,ringeval2019}, and gait~\cite{wang2021gait}; see
Song et al.~\cite{song2025survey} for a recent survey.
Semi-structured interviews remain the dominant elicitation
protocol~\cite{matsushima2024development}, the PHQ-9
questionnaire~\cite{kroenke2001} the standard severity measure, and
BiLSTMs with attention pooling remain competitive on small sequential
behavioural datasets~\cite{lin2020bilstm}.
The MPDD dataset~\cite{fu2026mpddavg,mpdd2026} uniquely adds free-text
personality descriptions as an explicit modality---relevant because
high Neuroticism and low Conscientiousness are among the strongest Big
Five predictors of depression risk~\cite{kotov2010}---but prior MPDD
systems (P3HF~\cite{p3hf}, Fosafer~\cite{fosafer}) measure the
personality contribution against \emph{soft}-ablation baselines that
retain the numerical Big Five scores.
On the encoder side, LLMs perform strongly on clinical
text~\cite{gatortron}, and LoRA~\cite{hu2022} enables domain
adaptation of 7B--72B models on small clinical datasets such as
DAIC-WOZ (51 subjects)~\cite{gratch2014}; however, fine-tuning
commonly degrades LLM calibration even when raw task performance
improves~\cite{zhu2023calibration,xiao2025restoring}, and DAIC-WOZ
adaptation risks learning disorder-general distress cues rather than
depression-specific markers~\cite{patapati2025daicwoz}---both directly
relevant to the failure mode we isolate in Section~\ref{sec:calib}.
Among the audio-visual features themselves, eGeMAPS~\cite{eyben2016}
provides 65 handcrafted features with established depression
biomarkers~\cite{williamson2016b} and OpenFace~\cite{baltrusaitis2018}
action units map onto depressive symptoms~\cite{girard2014}, a
clinical specificity that generic representations such as
wav2vec2~\cite{baevski2020} and DenseNet~\cite{huang2017densenet}
lack.
An extended review is given in Appendix~\ref{ssec:rw}.

\section{Task and Dataset}

\subsection{Task Formulation}
\label{sec:task}

Per subject, the challenge provides (by sub-track) gait IMU sequences,
audio-visual interview recordings, and a personality description
$\mathbf{d}$ (English text, $\approx$150 tokens) with structural
features $\mathbf{s}$ (Big Five scores, age, sex, financial stress,
family size, and disease class for Elder).
Interview audio (Mandarin) is transcribed with Whisper
large-v3~\cite{radford2023whisper} and encoded with
Chinese-RoBERTa~\cite{cui2021chinese} (768-dim [CLS]), mean-pooled
over a subject's 3--4 sessions into a transcript embedding
$\mathbf{t}$.
We predict \textbf{binary} status (depressed: PHQ-9 $\geq 10$),
\textbf{ternary} severity (none $<5$ / mild 5--10 / severe $\geq 10$),
and continuous \textbf{PHQ-9} $\hat{r}\in[0,27]$.

Following the official scoring rules~\cite{fu2026mpddavg,mpdd2026},
the Track-Level Score is $\text{TLS} = (\text{Cls\_F1} + \text{CCC} +
\kappa)/3$, where Cls\_F1 (macro-F1) and $\kappa$ each average their
binary and ternary counterparts and CCC is the PHQ-9 concordance
correlation coefficient~\cite{lin1989ccc}.
Tables additionally report \emph{ternary} Cohen's~$\kappa$ as an additional diagnostic, being more sensitive to chance-level
agreement on imbalanced severity classes.
The cross-track \emph{G-Score} is the mean TLS over all six
sub-tracks.

\subsection{Dataset and Sub-Tracks}
\label{sec:dataset}

Three modality sets---G+P (gait), A-V+P (audio-visual), A-V-G+P
(both), each with personality---are paired with the Elder and Young
cohorts.
\textbf{Elder}: 87 train, 23 test subjects; descriptions include a
disease class (neurological, endocrine, circulatory, other, healthy),
clinically relevant for late-life depression.
\textbf{Young}: 88 train, 22 test subjects; no disease class.
The cohorts' training PHQ-9 distributions have similar means (Young
4.8 vs.\ Elder 4.5) and severe counts (10 vs.\ 11), but Young is
\emph{compressed}: SD 3.5 vs.\ 4.7 with nearly twice as many
mild-band subjects (33 vs.\ 17), concentrated near the ternary
thresholds (Section~\ref{sec:cohorts}).

\paragraph{Transcript note.}
The transcript embedding $\mathbf{t}$ enters the personality stream
in \emph{all} sub-tracks, including G+P; the sub-track definitions
describe intended modality focus rather than an eligibility rule, and
the organisers' reproducibility verification reran our released
pipeline, transcript included.
The headline personality gain is not a transcript artefact: the
transcript is present on \emph{both} sides of the ablation, and L0
(gait$+$transcript, no personality) is at chance
(Section~\ref{sec:hier-results}).

\section{The HiPAD Architecture}

\subsection{Unified Variable-Modality Design}
\label{sec:arch}

\begin{figure*}[t]
  \centering
  \includegraphics[width=0.58\textwidth]{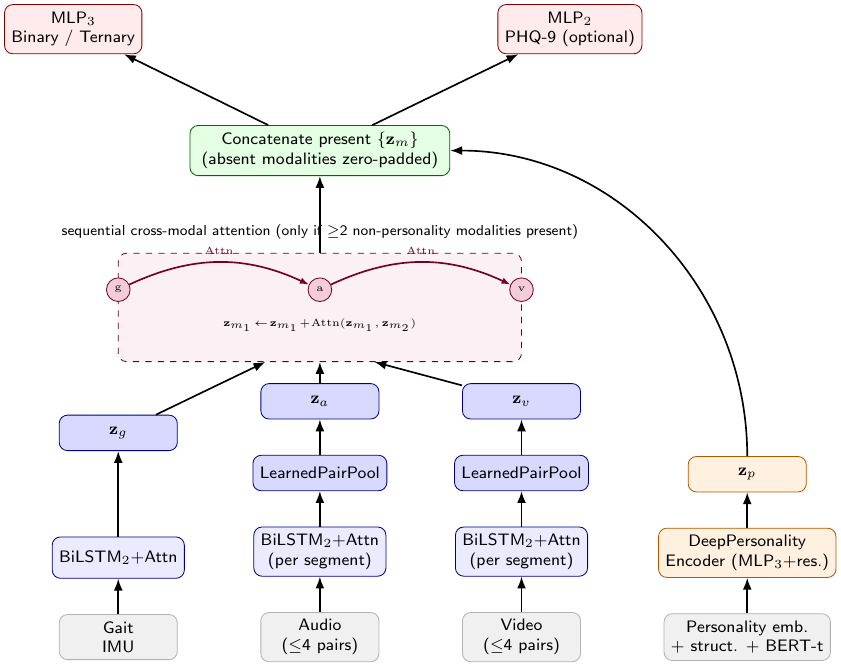}
  \caption{HiPAD pipeline. Gait, audio, and video pass through
  cross-modal attention \emph{only when at least two are present};
  personality always bypasses attention and joins at concatenation.
  Audio and video arrive as up to 4 interview segments, pooled by a
  learned softmax layer; absent modalities (per sub-track,
  Section~\ref{sec:dataset}) are zero-padded.}
  \label{fig:architecture}
\end{figure*}

\textbf{HiPAD} (\emph{Hierarchical Personality-Aware Depression}
model) is instantiated for each sub-track only through its modality
list (Section~\ref{sec:dataset}, Figure~\ref{fig:architecture};
feature matrix in Appendix~\ref{ssec:features}); in the released code
it is \texttt{EnhancedMPDDModel}.
Each available non-personality modality $m\in\{g,a,v\}$ (gait, audio,
video) is encoded by a 2-layer bidirectional LSTM with temporal
attention pooling and layer normalisation,
$\mathbf{z}_m = \mathrm{LN}\big(\textstyle\sum_t \alpha_t^{(m)}\,
\mathrm{BiLSTM}_2(\mathbf{f}_m)_t\big)$ with
$\alpha^{(m)} =
\mathrm{softmax}\big(W_a\,\mathrm{BiLSTM}_2(\mathbf{f}_m)\big)$,
where $\mathbf{f}_m\in\mathbb{R}^{T_m\times d_m}$ is the raw frame
sequence ($d_g$ gait, $d_a{\in}\{65,768\}$ audio, $d_v{=}710$ video),
linearly projected to the model width $h{=}128$.
$h$ is the \emph{final bidirectional} output width (each direction of
the 2-layer LSTM has hidden size $h/2{=}64$), so each pooled
$\mathbf{z}_m\in\mathbb{R}^{h}$ and the fused width below is
$h\cdot|\mathcal{M}|$; the temporal attention
$W_a\in\mathbb{R}^{1\times h}$ assigns one scalar weight per time step
(softmax over $t$).
Audio and video arrive as up to 4 independently-recorded
interview-segment ``pairs'' per subject (zero-padded with a binary
mask when fewer are available); each segment is encoded by the
\emph{same} BiLSTM and combined across segments by a second learned
softmax pooling layer (\emph{LearnedPairPool}).
The personality stream concatenates the personality embedding
$\mathbf{e}_p$ with small-MLP projections of the 18-dim structural
features $\mathbf{s}$ and the transcript $\mathbf{t}$, then passes
through a 3-layer MLP with a residual shortcut
(\emph{DeepPersonalityEncoder}):
\begin{equation}
  \label{eq:pers}
  \mathbf{z}_p = \mathrm{LN}\big(\mathrm{MLP}_3(\mathbf{x}) +
  W_{\text{res}}\mathbf{x}\big),\quad
  \mathbf{x} = [\mathbf{e}_p;\,\mathrm{MLP}_s(\mathbf{s});\,\mathrm{MLP}_t(\mathbf{t})],
\end{equation}
with $\mathbf{z}_p\in\mathbb{R}^{h}$ and $W_{\text{res}}$ a linear map
from the concatenated input to $\mathbb{R}^{h}$.

When two or more non-personality modalities are present (A-V+P,
A-V-G+P), each ordered pair $(m_1, m_2)$ updates its representation by
residual single-head scaled dot-product attention,
$\mathbf{z}_{m_1} \leftarrow \mathbf{z}_{m_1} +
W_o\,\mathrm{softmax}\!\big((W_q\mathbf{z}_{m_1})(W_k\mathbf{z}_{m_2})^{\!\top}/\sqrt{h}\big)\,
W_v\mathbf{z}_{m_2}$,
with $W_q,W_k,W_v,W_o\in\mathbb{R}^{h\times h}$.
Pairs are visited in the fixed order gait, audio, video and applied
sequentially; we did not tune this order.
For G+P, with a single non-personality modality, no cross-modal
attention is applied.
Personality is fused only by concatenation, never by cross-attention:
in this small-$n$ regime, letting the dominant personality signal
attend to the much weaker A-V streams degraded validation performance.
The per-modality vectors are concatenated
($\text{fused\_dim} = h\times|\mathcal{M}|$) and passed to a 3-layer
MLP classification head (binary or ternary, trained as separate models
per task and ensembled) and, when enabled, a parallel 2-layer MLP
regression head (PHQ-9).

\textbf{Features.}
Audio: OpenSMILE eGeMAPS v2 (65-dim) or wav2vec2-base (768-dim), per
Section~\ref{sec:audio}.
Video: OpenFace 2.0 (710-dim).
Gait: IMU sequences with stride, cadence, and asymmetry features.
Personality: RoBERTa-large~\cite{liu2019roberta} (1024-dim [CLS])
plus 18-dim structural features (Big Five~\cite{john1999big},
demographics, Elder disease class), present in all sub-tracks.
Full feature-extraction settings are in the supporting materials.

\subsection{Personality Encoding Hierarchy}
\label{sec:hierarchy}

We evaluate four levels of representation of the challenge's
\emph{personality-description modality}, whose free text bundles Big
Five traits with demographics and, for Elder, disease history.
We measure this modality's contribution as a whole; ``personality
gain'' throughout means \emph{personality-description-modality} gain,
without claiming demographics or disease class are personality
traits---disease class in particular is a direct clinical correlate
of late-life depression~\cite{hegeman2012}.

\begin{description}
\item[L0 --- No personality (true ablation).] Only the gait encoder
  and transcript remain; all personality-derived features are
  removed: no text embedding, no Big Five structural features, no
  demographic or disease flags.
\item[L1 --- Structural features only.] The description parsed into
  the 18-dim vector $\mathbf{s}$: Big Five (N, E, A, O, C
  $\in[1,15]$), age, sex, financial stress (0--4), family members,
  Elder disease class (one-hot). No text embedding.
\item[L2 --- Description embeddings.] Frozen RoBERTa-large (1024-dim
  [CLS]) plus $\mathbf{s}$.
\item[L3 --- Domain-adapted LLM.]
  \begin{sloppypar}
  \texttt{gte-Qwen2-7B-instruct}~\cite{alibaba2024} fine-tuned via
  LoRA ($r{=}64$, $\alpha{=}128$) in two stages---PHQ-8 regression on
  DAIC-WOZ~\cite{gratch2014}, then PHQ-9 regression on the MPDD train
  set (3584-dim); \texttt{Qwen2.5-72B-Instruct}~\cite{qwen2024} via
  4-bit QLoRA ($r{=}64$), same pipeline (8192-dim); and
  \texttt{e5-mistral-7b-instruct}~\cite{wang2023e5} (4096-dim, frozen)
  as a non-adapted control.
  \end{sloppypar}
\end{description}

\subsection{Training}
\label{sec:training}

Each task head is trained separately.
Binary and ternary heads use focal loss~\cite{lin2017}
($\gamma{=}2.0$, label smoothing 0.1); the ternary head adds an
ordinal expected-rank penalty ($\lambda_{\mathrm{ord}}{=}0.5$); the
auxiliary PHQ-9 head, when enabled, adds an MSE term
($\lambda_{\mathrm{reg}}{=}0.3$) on $\log(1{+}\cdot)$-normalised
scores.
Optimisation: AdamW, batch size 8, up to 120 epochs, early stopping
(patience 30 on validation macro-F1), weight decay $10^{-5}$ (G+P) /
$10^{-4}$ (A-V), MixUp~\cite{zhang2018mixup} ($\alpha{=}0.2$,
$p{=}0.5$), learning rate $8\times10^{-5}$ (G+P) / $3\times10^{-4}$
(A-V).
We use $k$-fold cross-validation ($k{=}4$ main, $k{=}8$ for the
ensemble-depth study) with 6 model seeds per fold (all seeds listed in
the supporting materials), ensembling the top-5 checkpoints by
validation $\kappa$.
Features are z-score-normalised per feature on training-split
statistics.
Ternary decoding (G+P): Elder uses the ternary head's argmax; Young
thresholds the model's own regression output $\hat r$ at 5.0/7.0
(thresholds on predictions, not labels).
Hardware details are given in the supporting materials.

\textbf{Reproducibility.}
Retrainable checkpoints and configurations for five of the six
sub-track configurations, the training script \texttt{train10.py},
and a one-command inference script for the final G+P configurations
are bundled in the released evaluation package; the encoder ablations
(L0, L1, L3) are retrainable from the released code and per-encoder
embeddings.
The supporting materials give the full feature-extraction and
optimisation settings, the CodaBench submission IDs for every number
reported here, the experiment-ID-to-checkpoint mapping, and the
per-run logs.

\section{Results}

\subsection{Cross-Track Scores}
\label{sec:results-lb}

\begin{table}[t]
\caption{Track-Level Scores at challenge close (2026-06-01);
Ter~$\kappa^{\ddagger}$ additional diagnostic, Rank$^{\dagger}$
interim standing (final ranking: Section~\ref{sec:limitations}).
The italicised Elder A-V+P row$^{\S}$ is from a separate, non-HiPAD
pipeline (Section~\ref{sec:avp-note}), excluded from all cross-track
comparisons; the unified G-Score uses the HiPAD value (0.397, only
the aggregate TLS available).}
\label{tab:results}
\footnotesize
\begin{tabular}{lccccc}
\toprule
Sub-track & TLS & Bin $\kappa$ & Ter $\kappa^{\ddagger}$ & CCC & Rank$^{\dagger}$ \\
\midrule
Elder G+P (16079)     & \textbf{0.647} & 0.572 & 0.677 & 0.549 & \#7/31 \\
Young G+P (16035)     & \textbf{0.578} & 0.636 & 0.405 & 0.527 & \#2 \\
Elder A-V+P (16078)   & 0.397 & --- & --- & --- & --- \\
Elder A-V-G+P (16077) & 0.477 & 0.330 & 0.319 & 0.504 & \#14/47 \\
Young A-V-G+P (16034) & 0.397 & 0.455 & 0.103 & 0.338 & \#2 \\
Young A-V+P (16033)   & 0.396 & 0.364 & 0.272 & 0.300 & \#2 \\
\midrule
\emph{Elder A-V+P, non-HiPAD}$^{\S}$ & \emph{0.484} & 0.439 & 0.434 & 0.374 & \#10/34 \\
\midrule
\textbf{G-Score (unified HiPAD)} & \textbf{0.482} & & & & \\
G-Score (best submitted$^{\S}$) & 0.496 & & & & \\
\bottomrule
\end{tabular}
\end{table}

G+P sub-tracks dominate (avg TLS 0.612 vs.\ 0.417 for A-V),
reflecting the decisive contribution of personality descriptions,
quantified next; on the organisers' reproducibility-verified final
leaderboard, our entries place 2nd on all three Young sub-tracks
(Section~\ref{sec:limitations}).

\paragraph{The Elder A-V+P entry.}
\label{sec:avp-note}
Our best-submitted Elder A-V+P predictions (TLS 0.484) came from a
separate pipeline developed independently by the second author
(\texttt{ReliabilityGatedEnhancedModel}: text-reliability gating with
personality-conditioned cross-attention), without $k$-fold
cross-validation and released only as precomputed predictions.
It is evidence neither about HiPAD nor about audio features, so we
exclude it from every cross-track comparison and headline claim; it
appears only in the separated rows of Tables~\ref{tab:results}
and~\ref{tab:audio}.
The retrainable HiPAD configuration (eGeMAPS$+$OpenFace) reaches
0.397 here; disentangling the $+0.087$ gap would require retraining
both pipelines under a common protocol, which the released artefacts
do not permit (Section~\ref{sec:limitations}).

\subsection{Hierarchical Personality Contribution}
\label{sec:hier-results}

We measure the personality-description-modality gain on the official
CodaBench leaderboard (Table~\ref{tab:hierarchy}) and isolate the
\emph{structured-vs-textual} split on held-out validation (full table in Appendix~\ref{ssec:hierarchy-val}), because the leaderboard closed before
a structured-only configuration could be scored on the test set.
All conditions use the identical architecture, $k{=}4$ folds, 6 seeds.

\begin{table}[t]
\caption{Personality-description-modality gain on the official
CodaBench leaderboard. L0: gait$+$transcript only, \emph{every}
personality-derived feature removed; L2: $+$18-dim structured
features and RoBERTa-large embedding; ``final'': best task-specific
ensemble.}
\label{tab:hierarchy}
\footnotesize
\begin{tabular}{clcccc}
\toprule
Level & Features & Elder & Young & $\Delta$E & $\Delta$Y \\
\midrule
L0 & Gait$+$transcript, no personality & 0.099 & 0.113 & --- & --- \\
L2 & $+$struct $+$ RoBERTa-large ($k{=}4$) & 0.632 & \textbf{0.578} & +0.533 & +0.465 \\
final & $+$task-specific ensemble & \textbf{0.647} & \textbf{0.578} & +0.548 & +0.465 \\
\bottomrule
\end{tabular}
\end{table}

The modality lifts TLS by \textbf{+0.499} on average at the
protocol-matched L2 level and \textbf{+0.507} with the final
ensembles; it also dwarfs every A-V score in
Table~\ref{tab:results}.
This gain is an order of magnitude larger than prior MPDD reports
(P3HF $+0.055$ accuracy~\cite{p3hf}, Fosafer $+0.029$--$0.057$
F1~\cite{fosafer}; metrics not directly comparable, table in Appendix~\ref{ssec:comparison}), for two structural reasons: soft ablations
retain the Big Five scores in the baseline, and in G+P gait alone is a
weak baseline, mechanically enlarging the \emph{relative}
contribution.
L0 models collapse to chance (binary $\kappa=0.00$, both cohorts):
gait$+$transcript alone carries no depression-severity signal in this
architecture, so the entire gain is attributable to the
personality-description modality (we diagnose the near-zero L0 itself
in Section~\ref{sec:true-soft}).

Structured Big Five and demographic features alone (L1) already
recover strong classification on validation (Elder binary
$\kappa{=}0.67$, Young $0.53$); adding the RoBERTa-large description
embedding (L2) improves Young markedly (binary
$\kappa\,0.53{\to}0.81$, ternary $0.41{\to}0.75$), while for Elder
the two are comparable on validation---an instance of the
validation-to-leaderboard gap analysed in Section~\ref{sec:capacity}.
We report these as relative validation comparisons because validation
overestimates held-out performance in this small-$n$ regime.

\subsection{Personality Encoder Comparison}
\label{sec:encoders}

Table~\ref{tab:encoder} compares personality encoders across the
hierarchy on the official leaderboard.

\begin{table}[t]
\caption{Encoder comparison on official CodaBench LB scores.
$^\dagger$val max $\kappa{=}0.91$ vs.\ LB 0.439 (overfitting;
Section~\ref{sec:capacity}). ``final'': binary $k{=}8$ $+$
ternary/PHQ $k{=}4$ ensembles.}
\label{tab:encoder}
\footnotesize
\begin{tabular}{lrcccr}
\toprule
Encoder & Dim & Bin $\kappa$ & Ter $\kappa^{\ddagger}$ & CCC & \textbf{TLS} \\
\midrule
\multicolumn{6}{l}{\emph{Elder G+P (16079)}}\\
None (L0)              & ---   & 0.000 & 0.082         & 0.030 & 0.099 \\
RoBERTa-large (L2)     & 1024  & 0.646 & \textbf{0.677}& 0.549 & 0.632 \\
LoRA 7B (L3a)          & 3584  & \textbf{0.776} & 0.553 & 0.415 & 0.582 \\
L3a$_{bin}$+BERT$_{ter}$ (L3b) & hyb & \textbf{0.776} & \textbf{0.677} & 0.420 & 0.624 \\
72B$_{QL}$+BERT$_{ter}$ (L3c) & hyb & 0.735 & \textbf{0.677} & 0.480 & 0.631 \\
QLoRA 72B full (L3d)   & 8192  & 0.735 & 0.480         & 0.358 & 0.543 \\
e5-mistral             & 4096  & $0.439^\dagger$ & 0.311 & 0.443 & 0.473 \\
Bin-8 + Tern/PHQ-4 (final) & hyb & 0.572 & \textbf{0.677}& \textbf{0.549} & \textbf{0.647} \\
\midrule
\multicolumn{6}{l}{\emph{Young G+P (16035)}}\\
None (L0)      & ---  & 0.000 & 0.000 & 0.006 & 0.113 \\
RoBERTa-large (L2) & 1024 & 0.636 & 0.405 & 0.527 & \textbf{0.578} \\
LoRA 7B (L3a)  & 3584 & 0.564 & 0.177 & 0.395 & 0.411 \\
\bottomrule
\end{tabular}
\end{table}

\subsection{Audio Feature Comparison}
\label{sec:audio}

\begin{table}[t]
\caption{Audio feature choice vs.\ A-V Track-Level Score at fixed
HiPAD architecture; the non-HiPAD Elder A-V+P entry
(Section~\ref{sec:avp-note}) below the rule is not a
same-architecture comparison.}
\label{tab:audio}
\footnotesize
\begin{tabular}{llccc}
\toprule
Sub-track & Cohort & wav2vec2 & eGeMAPS & Winner \\
\midrule
A-V+P (16078)   & Elder & --- & 0.397 & --- \\
A-V-G+P (16077) & Elder & 0.459 & \textbf{0.477} & eGeMAPS \\
A-V+P (16033)   & Young & 0.336 & \textbf{0.396} & eGeMAPS \\
A-V-G+P (16034) & Young & 0.386 & \textbf{0.397} & eGeMAPS \\
\midrule
\emph{A-V+P (16078), non-HiPAD}$^{\S}$ & Elder & \emph{0.484} & --- & --- \\
\bottomrule
\end{tabular}
\end{table}

\begin{figure}[t]
  \centering
  \includegraphics[width=0.62\columnwidth]{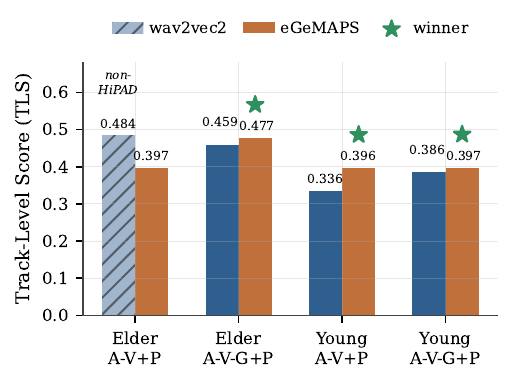}
  \caption{wav2vec2 vs.\ eGeMAPS Track-Level Score across the A-V
  sub-tracks (stars mark winners); the Elder A-V+P wav2vec2 bar is
  the non-HiPAD entry (Section~\ref{sec:avp-note}).}
  \label{fig:audio}
\end{figure}

eGeMAPS wins every same-architecture comparison
(Table~\ref{tab:audio}), and during development DenseNet video
features likewise underperformed OpenFace action units and were not
submitted.
The best Elder A-V+P entry nevertheless used wav2vec2 (0.484), but it
is the non-HiPAD pipeline of Section~\ref{sec:avp-note}, so its
margin says nothing about the audio representation; whether
self-supervised features capture elderly phonation cues that eGeMAPS
misses~\cite{martinez2025elderly} remains untested (extended discussion in Appendix~\ref{ssec:wav2vec-elder}).

\section{Analysis}

\subsection{LLM Calibration Mismatch}
\label{sec:calib}

Table~\ref{tab:encoder} reveals a dissociation (visualised in Appendix~\ref{ssec:calib}): LoRA 7B reaches binary $\kappa=0.776$
($+0.13$ over RoBERTa-large) yet a \emph{lower} TLS (0.582 vs.\
0.647).
TLS weights F1, $\kappa$, and CCC equally, so degraded regression or
ternary quality directly cancels binary gains.
The mechanism is calibration: LoRA 7B fine-tuned on PHQ regression
produces extreme continuous predictions (validation range 0.67--9.78
vs.\ 2.0--5.0 for RoBERTa-large), which (1) collapses CCC
($0.549{\to}0.415$) and (2) misfires the fixed ternary thresholds
(5.0/7.0) applied to out-of-range outputs (ternary $\kappa\,
0.677{\to}0.553$).
The hybrid L3b configuration (LoRA binary head + BERT ternary head)
confirms the diagnosis: restoring ternary quality raises TLS from
0.582 to 0.624 with binary unchanged.
Domain-adapted LLMs therefore require post-hoc PHQ-9 calibration
(isotonic regression or Platt scaling) before deployment in
TLS-optimised systems.
This extends the finding that fine-tuning degrades LLM
calibration~\cite{zhu2023calibration,xiao2025restoring} to a concrete
metric drop in clinical regression, compounded by the
DAIC-WOZ$\to$MPDD domain gap~\cite{patapati2025daicwoz}.

\subsection{Diagnosing the Chance-Level L0 Baseline}
\label{sec:true-soft}

Both cohorts' L0 models collapse to majority-class prediction (binary
$\kappa=0.00$), and two readings must be separated.
As an \emph{ablation baseline}, L0 shows the personality-description
modality---not gait---is the primary discriminative signal in G+P;
since a baseline retaining Big Five scores already reaches binary
$\kappa\approx0.53$--$0.67$ on validation
(Section~\ref{sec:hier-results}), soft designs substantially
understate the modality's contribution.
As a statement \emph{about gait}, L0 is only a lower bound for this
architecture class---gait detection on larger cohorts reports
meaningful discriminative power~\cite{wang2021gait}, and the challenge
provides no external no-personality baseline here---so we claim only
that a capacity-matched BiLSTM at $n\approx90$ recovers no
depression-severity signal from IMU gait plus transcript.

\subsection{Capacity Must Match $n{=}87$--$88$}
\label{sec:capacity}

Every change that added capacity failed to transfer to the test set,
whereas deeper ensembling, which adds no parameters, helped.
\textbf{Hidden width}: $h{=}256$ matched or slightly exceeded
$h{=}128$ on Elder validation ($\kappa$ 0.732 vs.\ $\sim$0.72), yet
all four submitted $h{=}256$ variants regressed on the leaderboard
($-0.004$ to $-0.038$ TLS); Young regressed on both.
\textbf{Embedding dimensionality}: e5-mistral (4096-dim) reached
validation binary $\kappa=0.91$ but the lowest submitted TLS (0.473,
Table~\ref{tab:encoder})---at $n{=}87$, a 4096-dim input memorises
validation splits, and unlike the regression-specific LoRA failure,
this degrades all three task heads.
\textbf{Ensemble depth} is the complementary lever, adding no
parameters: raising $k$ from 4 to 8 gained $+0.021$ TLS on Young, and
a task-specific hybrid combining the $k{=}8$ binary ensemble with the
$k{=}4$ ternary/PHQ ensemble gained $+0.015$ on Elder
($0.632{\to}0.647$); the hybrid, however, cherry-picks checkpoints
across two $k$-fold runs, which makes it sensitive to reproduction
variance---plausibly why it is absent from the organisers'
independently reproduced ranking (Section~\ref{sec:limitations}).
Full ensemble depths per configuration are in the supporting
materials.
We therefore use $h{=}128$ and $\leq$1024-dim embeddings uniformly,
which also enables cross-track analysis without confounding
architecture effects.
The same signature explains why \emph{adding modalities can hurt}:
the A-V-G+P tracks score below gait-only G+P (Elder
$0.647{\to}0.477$; Young $0.578{\to}0.397$) despite being a modality
superset, because concatenating two weak, noisy A-V streams beside a
dominant personality signal enlarges the fused width and raises
capacity without matching signal---diluting the decisive modality
exactly as extra hidden units do, and answering the apparent paradox
in Table~\ref{tab:results} (visualised as a radar chart in Appendix~\ref{ssec:radar}).

\subsection{Cohort Effects: Elder vs.\ Young}
\label{sec:cohorts}

Elder outperforms Young on every same-architecture comparison: G+P
$+0.069$ TLS, A-V-G+P $+0.080$, while A-V+P is effectively tied
(0.397 vs.\ 0.396, Table~\ref{tab:audio}).
The evidence localises the gap in severity grading, not detection:
Young ternary $\kappa$ is far below Elder's under identical models
(A-V+P: 0.272 vs.\ 0.434; G+P: 0.405 vs.\ 0.677), whereas Young
\emph{binary} $\kappa$ is the best we measured on G+P (0.636).
This matches the label distributions (Section~\ref{sec:dataset}):
Young's PHQ-9 scores are compressed (SD 3.5 vs.\ 4.7) with twice the
mass in the mild band, i.e., concentrated near the ternary
thresholds, so equal-quality predictions yield more boundary errors.
It is also consistent with---though not direct evidence for---older
adults presenting more somatic and psychomotor symptoms, younger
adults more cognitive ones~\cite{hegeman2012}.
The same asymmetry recurs inside the personality modality (Elder's
gain +0.548 vs.\ Young's +0.465, partly because Elder descriptions
include disease class, a direct clinical correlate) and in gait:
holding HiPAD/eGeMAPS fixed, adding gait to A-V+P gains $+0.080$ TLS
for Elder but $+0.001$ for Young (Table~\ref{tab:audio})---again the
psychomotor-channel pattern; and domain-adapted LLMs hurt Young more
(ternary $\kappa\,0.405{\to}0.177$), consistent with the DAIC-WOZ
fine-tuning distribution diverging further from Young's compressed
PHQ profile.
Together this suggests Young-cohort progress needs richer linguistic
features or finer-grained severity annotation, not further
architecture changes.

\section{Discussion}
\label{sec:discussion}

The personality ablation and the system results describe the same
effect from two sides: on the gait tracks, L0 shows that personality
is essentially the only informative modality, and on the audio-visual
tracks the same imbalance explains why adding modalities can lower
the score (Section~\ref{sec:capacity}).

Three of our observations extend prior work rather than repeat it: the
personality gain of 0.499 to 0.507 shows that the small soft-ablation
gains reported by P3HF and Fosafer are mostly an artefact of leaving
the Big Five scores in the baseline (Section~\ref{sec:hier-results});
the calibration failure of the adapted LLMs is a clinical-regression
instance of fine-tuning degrading LLM
calibration~\cite{zhu2023calibration,xiao2025restoring}, with the
hybrid-head experiment pinning the damage to the regression output;
and eGeMAPS and OpenFace beating generic deep features at $n{<}100$
carries the AVEC-era experience with handcrafted paralinguistic
features~\cite{ringeval2019,williamson2016b} over to a new dataset
and cohort split.

For practitioners with small clinical cohorts: a frozen mid-sized
encoder with deeper ensembling is safer than a domain-adapted LLM
unless the regression output is re-calibrated after adaptation (e.g.,
isotonic regression or Platt scaling), and a result from a different
model class should not be used to draw conclusions about features
(Section~\ref{sec:audio}). Promising directions are
reliability-weighted fusion tolerating missing modalities,
personality-conditioned attention, and adaptation to non-Chinese
populations. On the evaluation side, per-component score reporting
(F1, CCC, $\kappa$) and a challenge-defined true-ablation protocol
would make calibration failures visible and personality-gain numbers
comparable across teams.

\section{Limitations}
\label{sec:limitations}

Each cohort provides only $n{=}87$--$88$ training and 22--23 test
subjects: this forces $h{=}128$, makes cross-validation noisy, and
lets a handful of misclassified subjects shift leaderboard positions;
we therefore report no confidence intervals or significance tests,
and controlled module ablations (cross-modal attention,
LearnedPairPool, residual personality encoder, single-model
vs.\ ensemble) remain future work.
The unified-architecture claim holds for five of six sub-tracks; the
sixth is the non-HiPAD Elder A-V+P entry (Section~\ref{sec:avp-note}),
released only as precomputed predictions rather than retrainable
weights.

\paragraph{Final ranking.}
The interim CodaBench standings at challenge close were superseded by
the organisers' reproducibility-verified final leaderboard, which we
treat as authoritative: it confirms rank~2nd on all three Young
sub-tracks; our Elder entries do not appear in its published top-3, so
Elder scores in Table~\ref{tab:results} are our CodaBench scores at
challenge close.
All submissions were finalised before the deadline and are
reproducible from the released checkpoints, configurations, and logs.

The clinical population is exclusively Chinese (Mandarin interviews,
English personality descriptions), so generalisability to other
languages and populations is untested; and only the aggregate TLS was
exposed per submission, limiting post-hoc error analysis.

\paragraph{Ethical use.}
This system detects statistical correlates of depression from
personality, speech, face, and gait---sensitive health data---and is
intended only as \emph{screening support} to prioritise clinical
attention, never as standalone diagnosis; deployment would require
consent, clinician oversight, bias auditing across demographic
groups, and validation beyond the single cohort studied here.

\section{Conclusion}

We submitted one fixed BiLSTM fusion network to all six MPDD-AVG 2026
sub-tracks, comparing cohorts and modality sets without architecture
confounds.
The one result a reader should take away is how much the personality
description matters: 150 tokens of text about who the subject is
predict depression better than audio, video, and gait combined, and a
model with gait and personality alone (TLS 0.647) beats every
configuration that adds audio and video; a true ablation puts the
contribution at 0.499--0.507 TLS, and without it the model is at
chance.
Domain-adapting the personality encoder improves classification but
distorts the PHQ-9 regression, so a frozen RoBERTa-large encoder with
deeper ensembling is the better choice at this scale; handcrafted
audio-visual features beat generic deep features in every
fixed-architecture comparison; and larger hidden widths, larger
embeddings, and extra modalities all overfit alike on fewer than 100
subjects.
The main obstacles at this sample size are representation size and
output calibration rather than architecture---PHQ-9 calibration in
particular deserves more attention in multi-task depression
assessment.

\begin{acks}
The authors thank the MPDD-AVG 2026 organisers for the dataset and
evaluation infrastructure, and \L ukasz Latkowski for his support
during the challenge, his initial work on transcriptions, and access
to his server and NVIDIA RTX~3090.
\end{acks}

\appendix

\section*{Appendix}
\noindent The material below was omitted from the conference version to meet
its page limit and is restored here; it was submitted with the paper as
supplementary material and is referenced from the main text above.

\section{Extended Related Work}
\label{ssec:rw}

\subsection{Multimodal Depression Detection}

Depression detection has been studied using audio features
(MFCCs, eGeMAPS, deep spectral
representations)~\cite{eyben2016,chlasta2019speech,chlasta2021spontaneous},
video features (action units, head pose)~\cite{baltrusaitis2018}, text
transcripts~\cite{depresym}, audio-visual shared
tasks~\cite{avec2016}, and gait~\cite{wang2021gait}; see Song et
al.~\cite{song2025survey} for a recent survey.
Semi-structured interviews remain the dominant elicitation
protocol~\cite{matsushima2024development}, and the PHQ-9
questionnaire~\cite{kroenke2001} the standard severity measure.
Multimodal fusion typically outperforms unimodal approaches, with
feature-level concatenation a strong baseline~\cite{ringeval2019};
BiLSTMs with attention pooling remain competitive on small sequential
behavioural datasets~\cite{lin2020bilstm}.
The MPDD dataset~\cite{fu2026mpddavg,mpdd2026} adds personality
descriptions as an explicit modality alongside audio-visual interviews
and IMU-based gait.

\subsection{Personality in Depression Detection}

High Neuroticism and low Conscientiousness are among the strongest
Big Five predictors of depression risk~\cite{kotov2010}.
MPDD uniquely provides per-subject personality descriptions in natural
language.
Prior MPDD systems~\cite{p3hf,fosafer} report personality gains against
baselines that retain Big Five numerical values; to our knowledge, no
prior work performs a fully personality-agnostic ablation within the
MPDD framework.

\subsection{LLM Embeddings for Clinical NLP}

LLMs perform strongly on clinical text~\cite{gatortron}, and
LoRA~\cite{hu2022} enables domain adaptation of 7B--72B models on small
clinical datasets such as DAIC-WOZ (51 subjects)~\cite{gratch2014}.
However, fine-tuning and alignment commonly degrade LLM calibration
even when raw task performance
improves~\cite{zhu2023calibration,xiao2025restoring}, and DAIC-WOZ
adaptation risks learning disorder-general distress cues rather than
depression-specific markers~\cite{patapati2025daicwoz}---both directly
relevant to the failure mode isolated in Section~6.1 (main paper).

\subsection{Audio and Video Features for Depression}

eGeMAPS~\cite{eyben2016} provides 65 handcrafted prosodic, spectral,
and cepstral features whose depression biomarkers include reduced F0
range, decreased loudness variance, and increased
jitter/shimmer~\cite{williamson2016b}; wav2vec2~\cite{baevski2020}
provides 768-dim self-supervised speech representations without
clinical priors.
On video, OpenFace 2.0~\cite{baltrusaitis2018} action units map onto
depression symptoms (AU6: flat affect; AU45: psychomotor retardation;
AU15)~\cite{girard2014}, a clinical specificity generic CNN features
such as DenseNet~\cite{huang2017densenet} lack.

\section{Input Features by Sub-Track}
\label{ssec:features}

Table~\ref{stab:features} gives the feature matrix referenced in
Section~4.1 (main paper). The transcript embedding is the
Chinese-RoBERTa encoding of the Whisper-transcribed interview audio,
fused into the personality stream in \emph{all} sub-tracks.

\begin{table}[h]
\caption{Input features by sub-track. \checkmark = present,
  --- = absent (zero-padded).}
\label{stab:features}
\small
\begin{tabular}{lccccc}
\toprule
Sub-track & Gait & Audio & Video & Personality & Transcript \\
\midrule
G+P       & \checkmark & ---        & ---        & \checkmark & \checkmark \\
A-V+P     & ---        & \checkmark & \checkmark & \checkmark & \checkmark \\
A-V-G+P   & \checkmark & \checkmark & \checkmark & \checkmark & \checkmark \\
\bottomrule
\end{tabular}
\end{table}

Full feature detail: audio is OpenSMILE eGeMAPS v2 (65-dim, 100\,ms
frame shift) or wav2vec2-base (768-dim); video is OpenFace 2.0
(710-dim: AU intensities/presences, 3-D head pose, eye gaze); gait is
IMU sequences (accelerometer $+$ gyroscope) with stride, cadence, and
asymmetry features; personality is RoBERTa-large (1024-dim [CLS])
plus the 18-dim structural features (Big Five, demographics, Elder
disease class).

\section{Cross-Track G-Score Radar Chart}
\label{ssec:radar}

\begin{figure}[h]
  \centering
  \includegraphics[width=0.8\columnwidth]{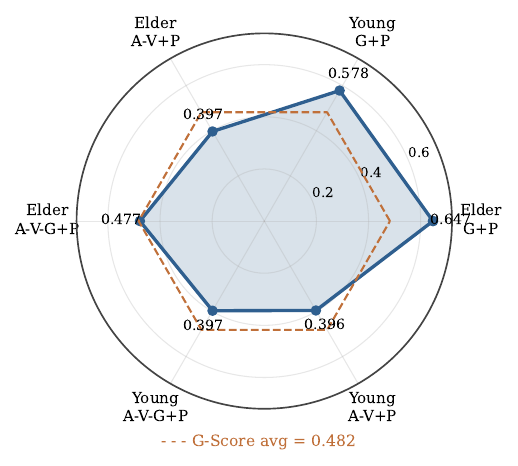}
  \caption{Track-Level Score by sub-track; the dashed line marks the
  unified-HiPAD cross-track G-Score (0.482). G+P sub-tracks visibly
  dominate the A-V sub-tracks.}
  \label{sfig:radar}
\end{figure}

Figure~\ref{sfig:radar} visualises the Track-Level Scores of Table~1
(main paper).

\section{Structured-vs-Textual Split on Held-Out Validation}
\label{ssec:hierarchy-val}

Table~\ref{stab:hierarchy-val} is the full validation table
summarised in Section~5.2 (main paper).

\begin{table}[h]
\caption{Structured-vs-textual split on held-out validation (top-5
ensemble by validation $\kappa$, mean over folds/seeds;
\emph{post-challenge, validation only}). L1: 18-dim structured
features only; L2 adds the RoBERTa-large description embedding.}
\label{stab:hierarchy-val}
\small
\begin{tabular}{lcccc}
\toprule
 & \multicolumn{2}{c}{Elder} & \multicolumn{2}{c}{Young} \\
Level & Bin $\kappa$ & Ter $\kappa$ & Bin $\kappa$ & Ter $\kappa$ \\
\midrule
L1 (struct only)     & 0.669 & 0.495 & 0.527 & 0.411 \\
L2 ($+$RoBERTa-large) & 0.554 & 0.559 & \textbf{0.814} & \textbf{0.751} \\
\bottomrule
\end{tabular}
\end{table}

\begin{figure}[h]
  \centering
  \includegraphics[width=0.8\columnwidth]{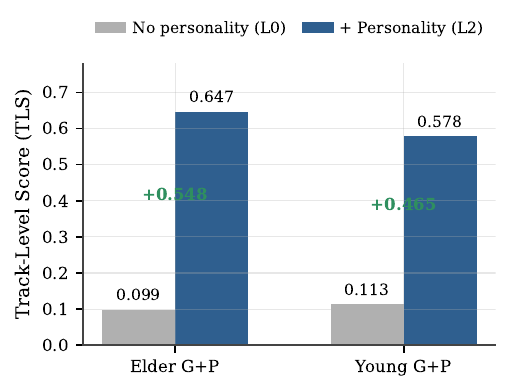}
  \caption{Track-Level Score with and without personality (true
  ablation), Elder and Young G+P. The gain is far larger than prior
  soft-ablation reports, though measured on a different metric and
  ablation definition (Table~\ref{stab:comparison}).}
  \label{sfig:gain}
\end{figure}

Figure~\ref{sfig:gain} shows the corresponding leaderboard-level
personality gain (true ablation) for both cohorts.

\section{Comparison with Prior MPDD Personality Ablations}
\label{ssec:comparison}

Table~\ref{stab:comparison} contrasts our ablation design with prior
MPDD systems, as summarised in Section~5.2 (main paper).

\begin{table}[h]
\caption{Personality gain in prior MPDD systems vs.\ ours. The rows
use \emph{different} metrics and ablation definitions and are
\textbf{not directly comparable}; the table contrasts ablation design,
not a common scale.}
\label{stab:comparison}
\footnotesize
\begin{tabular}{lllcr}
\toprule
System & Encoder & Ablation & Metric & Gain \\
\midrule
P3HF~\cite{p3hf}    & GPT-4$\to$BERT-768   & Soft & Acc.\ & +0.055 \\
Fosafer~\cite{fosafer} & ChatGLM3$\to$RoBERTa & Soft & F1   & +0.029--0.057 \\
\textbf{WarsawIQ}   & \textbf{RoBERTa-l.} & \textbf{True} & \textbf{TLS} & \textbf{+0.499} \\
\bottomrule
\end{tabular}
\end{table}

The substantially larger gain has two structural sources: our ablation
removes \emph{all} personality-derived information, whereas soft
ablations retain Big Five scores; and in the G+P setting gait alone is
a weak baseline, mechanically enlarging personality's \emph{relative}
contribution.

\section{LLM Calibration Scatter}
\label{ssec:calib}

\begin{figure}[h]
  \centering
  \includegraphics[width=0.8\columnwidth]{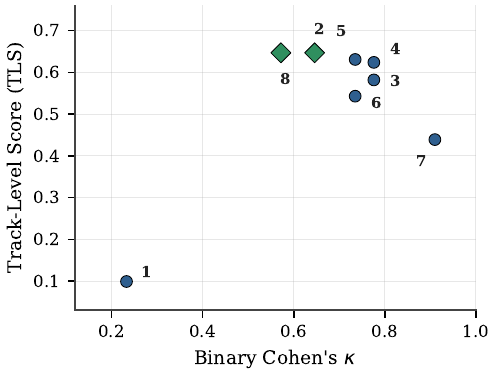}
  \caption{Binary $\kappa$ vs.\ TLS for all Elder G+P personality
  encoders (Table~4, main paper): 1 None (L0), 2 RoBERTa-large
  (L2), 3 LoRA~7B (L3a), 4 L3a~bin+BERT~ter, 5 QLoRA72B+BERT~ter,
  6 QLoRA72B~full, 7 e5-mistral, 8 final. Diamonds (2, 8) reach the
  best TLS despite very different binary $\kappa$.}
  \label{sfig:calib}
\end{figure}

Figure~\ref{sfig:calib} visualises the dissociation between binary
$\kappa$ and TLS analysed in Section~6.1 (main paper).

\section{Ensemble Depth}
\label{ssec:ensemble}

Deeper ensembling is the one capacity lever that helps at this $n$
(Section~6.3, main paper): raising $k$ from 4 to 8 gains $+0.021$
TLS on Young (unified), and a task-specific hybrid combining the
$k{=}8$ binary ensemble with the $k{=}4$ ternary/PHQ ensemble gains
$+0.015$ on Elder (0.632$\to$0.647); full depths per configuration are
in the supporting-materials logs.
The hybrid, however, cherry-picks checkpoints per task across two
$k$-fold runs, making it more sensitive to reproduction variance than
a single deterministic pipeline---plausibly why it is absent from the
organisers' independently reproduced ranking.

\section{Is wav2vec2 Better for Elder Speech?}
\label{ssec:wav2vec-elder}

A natural question is why the best Elder A-V+P entry used wav2vec2 (0.484)
while wav2vec2 \emph{loses} the same-architecture Elder A-V-G+P
comparison (0.459 vs.\ 0.477, Table~5, main paper).
The two numbers come from different model classes: the 0.484 entry is
the non-HiPAD pipeline of Section~5.1 (main paper), so its margin
says nothing about the audio representation; within HiPAD, eGeMAPS
wins every comparison.
The hypothesis that self-supervised representations capture
phonation-level cues of elderly depressed speech (tremor, breathiness,
creakiness) inadequately summarised by eGeMAPS's 65
statistics~\cite{martinez2025elderly} remains plausible but untested:
the controlled same-architecture Elder A-V+P retraining is the
experiment our released artefacts do not currently support.

\section{Training Details}
\label{ssec:training}

\textbf{Seeds and folds.} $k$-fold cross-validation uses fold seed
42; the 6 model seeds per fold are $\{42, 128, 256, 512, 789,
1024\}$.

\textbf{Loss.} The combined objective is
$\mathcal{L} = \mathcal{L}_{\mathrm{focal}} +
\lambda_{\mathrm{ord}}\mathcal{L}_{\mathrm{ord}} +
\lambda_{\mathrm{reg}}\mathcal{L}_{\mathrm{reg}}$ with
$\lambda_{\mathrm{ord}}{=}0.5$ (0 for binary) and
$\lambda_{\mathrm{reg}}{=}0.3$, as specified in Section~4.3 (main
paper).

\textbf{Ternary decoding (G+P).} Young's regression-output thresholds
(5.0/7.0) are applied to predictions, not labels, and were tuned to
the cohort's score distribution.

\textbf{Hardware.} Training ran on an AMD Radeon 890M (ROCm 6.3.1),
an NVIDIA RTX~3090, and an A100 SXM4 80\,GB (QLoRA 72B only).

\bibliographystyle{ACM-Reference-Format}
\balance
\bibliography{unified}


\begin{thebibliography}{40}


\ifx \showCODEN    \undefined \def \showCODEN     #1{\unskip}     \fi
\ifx \showISBNx    \undefined \def \showISBNx     #1{\unskip}     \fi
\ifx \showISBNxiii \undefined \def \showISBNxiii  #1{\unskip}     \fi
\ifx \showISSN     \undefined \def \showISSN      #1{\unskip}     \fi
\ifx \showLCCN     \undefined \def \showLCCN      #1{\unskip}     \fi
\ifx \shownote     \undefined \def \shownote      #1{#1}          \fi
\ifx \showarticletitle \undefined \def \showarticletitle #1{#1}   \fi
\ifx \showURL      \undefined \def \showURL       {\relax}        \fi
\providecommand\bibfield[2]{#2}
\providecommand\bibinfo[2]{#2}
\providecommand\natexlab[1]{#1}
\providecommand\showeprint[2][]{arXiv:#2}

\bibitem[{Alibaba-NLP}(2024)]%
        {alibaba2024}
\bibfield{author}{\bibinfo{person}{{Alibaba-NLP}}.}
  \bibinfo{year}{2024}\natexlab{}.
\newblock \bibinfo{title}{{gte-Qwen2-7B-instruct}}.
\newblock \bibinfo{howpublished}{HuggingFace Model Hub.
  \url{https://huggingface.co/Alibaba-NLP/gte-Qwen2-7B-instruct}}.
\newblock


\bibitem[Baevski et~al\mbox{.}(2020)]%
        {baevski2020}
\bibfield{author}{\bibinfo{person}{Alexei Baevski}, \bibinfo{person}{Yuhao
  Zhou}, \bibinfo{person}{Abdelrahman Mohamed}, {and} \bibinfo{person}{Michael
  Auli}.} \bibinfo{year}{2020}\natexlab{}.
\newblock \showarticletitle{wav2vec 2.0: A Framework for Self-Supervised
  Learning of Speech Representations}. In \bibinfo{booktitle}{\emph{Advances in
  Neural Information Processing Systems (NeurIPS)}}, Vol.~\bibinfo{volume}{33}.
  \bibinfo{pages}{12449--12460}.
\newblock


\bibitem[Baltru\v{s}aitis et~al\mbox{.}(2018)]%
        {baltrusaitis2018}
\bibfield{author}{\bibinfo{person}{Tadas Baltru\v{s}aitis},
  \bibinfo{person}{Amir Zadeh}, \bibinfo{person}{Yao~Chong Lim}, {and}
  \bibinfo{person}{Louis-Philippe Morency}.} \bibinfo{year}{2018}\natexlab{}.
\newblock \showarticletitle{{OpenFace} 2.0: Facial Behaviour Analysis Toolkit}.
  In \bibinfo{booktitle}{\emph{2018 13th IEEE International Conference on
  Automatic Face \& Gesture Recognition (FG)}}. \bibinfo{pages}{59--66}.
\newblock
\href{https://doi.org/10.1109/FG.2018.00019}{doi:\nolinkurl{10.1109/FG.2018.00019}}


\bibitem[Chlasta et~al\mbox{.}(2021)]%
        {chlasta2021spontaneous}
\bibfield{author}{\bibinfo{person}{Karol Chlasta}, \bibinfo{person}{Pawe{\l}
  Holas}, {and} \bibinfo{person}{Krzysztof Wo{\l}k}.}
  \bibinfo{year}{2021}\natexlab{}.
\newblock \showarticletitle{Computer-Based Detection of Depression and Dementia
  in Spontaneous Speech}.
\newblock \bibinfo{journal}{\emph{European Psychiatry}} \bibinfo{volume}{64},
  \bibinfo{number}{S1} (\bibinfo{year}{2021}), \bibinfo{pages}{S349--S349}.
\newblock
\href{https://doi.org/10.1192/j.eurpsy.2021.936}{doi:\nolinkurl{10.1192/j.eurpsy.2021.936}}


\bibitem[Chlasta et~al\mbox{.}(2019)]%
        {chlasta2019speech}
\bibfield{author}{\bibinfo{person}{Karol Chlasta}, \bibinfo{person}{Krzysztof
  Wo{\l}k}, {and} \bibinfo{person}{Izabela Krejtz}.}
  \bibinfo{year}{2019}\natexlab{}.
\newblock \showarticletitle{Automated Speech-Based Screening of Depression
  Using Deep Convolutional Neural Networks}.
\newblock \bibinfo{journal}{\emph{Procedia Computer Science}}
  \bibinfo{volume}{164} (\bibinfo{year}{2019}), \bibinfo{pages}{618--628}.
\newblock
\href{https://doi.org/10.1016/j.procs.2019.12.228}{doi:\nolinkurl{10.1016/j.procs.2019.12.228}}


\bibitem[Cui et~al\mbox{.}(2021)]%
        {cui2021chinese}
\bibfield{author}{\bibinfo{person}{Yiming Cui}, \bibinfo{person}{Wanxiang Che},
  \bibinfo{person}{Ting Liu}, \bibinfo{person}{Bing Qin}, {and}
  \bibinfo{person}{Ziqing Yang}.} \bibinfo{year}{2021}\natexlab{}.
\newblock \showarticletitle{Pre-Training with Whole Word Masking for {Chinese}
  {BERT}}.
\newblock \bibinfo{journal}{\emph{IEEE/ACM Transactions on Audio, Speech, and
  Language Processing}}  \bibinfo{volume}{29} (\bibinfo{year}{2021}),
  \bibinfo{pages}{3504--3514}.
\newblock
\href{https://doi.org/10.1109/TASLP.2021.3124365}{doi:\nolinkurl{10.1109/TASLP.2021.3124365}}


\bibitem[Eyben et~al\mbox{.}(2016)]%
        {eyben2016}
\bibfield{author}{\bibinfo{person}{Florian Eyben}, \bibinfo{person}{Klaus~R.
  Scherer}, \bibinfo{person}{Björn~W. Schuller}, \bibinfo{person}{Johan
  Sundberg}, \bibinfo{person}{Elisabeth Andr\'{e}}, \bibinfo{person}{Carlos
  Busso}, \bibinfo{person}{Laurence~Y. Devillers}, \bibinfo{person}{Julien
  Epps}, \bibinfo{person}{Petri Laukka}, \bibinfo{person}{Shrikanth~S.
  Narayanan}, {and} \bibinfo{person}{Khiet~P. Truong}.}
  \bibinfo{year}{2016}\natexlab{}.
\newblock \showarticletitle{The {Geneva Minimalistic Acoustic Parameter Set
  (GeMAPS)} for Voice Research and Affective Computing}.
\newblock \bibinfo{journal}{\emph{IEEE Transactions on Affective Computing}}
  \bibinfo{volume}{7}, \bibinfo{number}{2} (\bibinfo{year}{2016}),
  \bibinfo{pages}{190--202}.
\newblock
\href{https://doi.org/10.1109/TAFFC.2015.2457417}{doi:\nolinkurl{10.1109/TAFFC.2015.2457417}}


\bibitem[Fu et~al\mbox{.}(2025a)]%
        {mpdd2026}
\bibfield{author}{\bibinfo{person}{Changzeng Fu}, \bibinfo{person}{Zelin Fu},
  \bibinfo{person}{Qi Zhang}, \bibinfo{person}{Xinhe Kuang},
  \bibinfo{person}{Jiacheng Dong}, \bibinfo{person}{Kaifeng Su},
  \bibinfo{person}{Yikai Su}, \bibinfo{person}{Wenbo Shi},
  \bibinfo{person}{Junfeng Yao}, \bibinfo{person}{Yuliang Zhao},
  \bibinfo{person}{Shiqi Zhao}, \bibinfo{person}{Jiadong Wang},
  \bibinfo{person}{Siyang Song}, \bibinfo{person}{Chaoran Liu},
  \bibinfo{person}{Yuichiro Yoshikawa}, \bibinfo{person}{Björn Schuller},
  {and} \bibinfo{person}{Hiroshi Ishiguro}.} \bibinfo{year}{2025}\natexlab{a}.
\newblock \showarticletitle{The First {MPDD} Challenge: Multimodal
  Personality-Aware Depression Detection}. In
  \bibinfo{booktitle}{\emph{Proceedings of the 33rd ACM International
  Conference on Multimedia (MM '25)}}. \bibinfo{pages}{13924--13929}.
\newblock
\href{https://doi.org/10.1145/3746027.3762020}{doi:\nolinkurl{10.1145/3746027.3762020}}


\bibitem[Fu et~al\mbox{.}(2026)]%
        {fu2026mpddavg}
\bibfield{author}{\bibinfo{person}{Changzeng Fu}, \bibinfo{person}{Yiming
  Zhang}, \bibinfo{person}{Ming Shang}, \bibinfo{person}{Shiqi Zhao},
  \bibinfo{person}{Alexis Meneses}, \bibinfo{person}{Zhaojie Luo},
  \bibinfo{person}{Chaoran Liu}, \bibinfo{person}{Wenbo Shi},
  \bibinfo{person}{Jiadong Wang}, \bibinfo{person}{Yuichiro Yoshikawa},
  \bibinfo{person}{Hiroshi Ishiguro}, {and} \bibinfo{person}{Björn Schuller}.}
  \bibinfo{year}{2026}\natexlab{}.
\newblock \bibinfo{title}{{MPDD-AVG}: Multimodal Personality-Aware Depression
  Detection via Audio-Visual Interview and Gait Analysis}.
\newblock
\shownote{MPDD-AVG Challenge 2026 dataset/system paper}.
\newblock


\bibitem[Fu et~al\mbox{.}(2025b)]%
        {p3hf}
\bibfield{author}{\bibinfo{person}{Changzeng Fu}, \bibinfo{person}{Shiwen
  Zhao}, \bibinfo{person}{Yunze Zhang}, \bibinfo{person}{Zhongquan Jian},
  \bibinfo{person}{Shiqi Zhao}, {and} \bibinfo{person}{Chaoran Liu}.}
  \bibinfo{year}{2025}\natexlab{b}.
\newblock \bibinfo{title}{Personality-Guided Public-Private Domain Disentangled
  Hypergraph-Former Network for Multimodal Depression Detection}.
\newblock
\shownote{Accepted at AAAI 2026}.
\newblock
\showeprint[arxiv]{2511.12460}~[cs.LG]


\bibitem[Girard et~al\mbox{.}(2014)]%
        {girard2014}
\bibfield{author}{\bibinfo{person}{Jeffrey~M. Girard},
  \bibinfo{person}{Jeffrey~F. Cohn}, \bibinfo{person}{Mohammad~H. Mahoor},
  \bibinfo{person}{S.~Mohammad Mavadati}, \bibinfo{person}{Zakia Hammal}, {and}
  \bibinfo{person}{David~P. Rosenwald}.} \bibinfo{year}{2014}\natexlab{}.
\newblock \showarticletitle{Nonverbal Social Withdrawal in Depression: Evidence
  from Manual and Automatic Analyses}.
\newblock \bibinfo{journal}{\emph{Image and Vision Computing}}
  \bibinfo{volume}{32}, \bibinfo{number}{10} (\bibinfo{year}{2014}),
  \bibinfo{pages}{641--647}.
\newblock
\href{https://doi.org/10.1016/j.imavis.2013.12.007}{doi:\nolinkurl{10.1016/j.imavis.2013.12.007}}


\bibitem[Gratch et~al\mbox{.}(2014)]%
        {gratch2014}
\bibfield{author}{\bibinfo{person}{Jonathan Gratch}, \bibinfo{person}{Ron
  Artstein}, \bibinfo{person}{Gale Lucas}, \bibinfo{person}{Giota Stratou},
  \bibinfo{person}{Stefan Scherer}, \bibinfo{person}{Angela Nazarian},
  \bibinfo{person}{Rachel Wood}, \bibinfo{person}{Jill Boberg},
  \bibinfo{person}{David DeVault}, \bibinfo{person}{Stacy Marsella},
  \bibinfo{person}{David~R. Traum}, \bibinfo{person}{Skip Rizzo}, {and}
  \bibinfo{person}{Louis-Philippe Morency}.} \bibinfo{year}{2014}\natexlab{}.
\newblock \showarticletitle{The Distress Analysis Interview Corpus of Human and
  Computer Interviews}. In \bibinfo{booktitle}{\emph{Proceedings of the Ninth
  International Conference on Language Resources and Evaluation (LREC)}}.
  \bibinfo{pages}{3123--3128}.
\newblock


\bibitem[Hegeman et~al\mbox{.}(2012)]%
        {hegeman2012}
\bibfield{author}{\bibinfo{person}{Johanna~M. Hegeman}, \bibinfo{person}{Rob~M.
  Kok}, \bibinfo{person}{Roos~C. van~der Mast}, {and} \bibinfo{person}{Erik~J.
  Giltay}.} \bibinfo{year}{2012}\natexlab{}.
\newblock \showarticletitle{Phenomenology of Depression in Older Compared with
  Younger Adults: Meta-Analysis}.
\newblock \bibinfo{journal}{\emph{British Journal of Psychiatry}}
  \bibinfo{volume}{200}, \bibinfo{number}{4} (\bibinfo{year}{2012}),
  \bibinfo{pages}{275--281}.
\newblock
\href{https://doi.org/10.1192/bjp.bp.111.095950}{doi:\nolinkurl{10.1192/bjp.bp.111.095950}}


\bibitem[Hu et~al\mbox{.}(2022)]%
        {hu2022}
\bibfield{author}{\bibinfo{person}{Edward~J. Hu}, \bibinfo{person}{Yelong
  Shen}, \bibinfo{person}{Phillip Wallis}, \bibinfo{person}{Zeyuan Allen-Zhu},
  \bibinfo{person}{Yuanzhi Li}, \bibinfo{person}{Shean Wang},
  \bibinfo{person}{Lu Wang}, {and} \bibinfo{person}{Weizhu Chen}.}
  \bibinfo{year}{2022}\natexlab{}.
\newblock \showarticletitle{{LoRA}: Low-Rank Adaptation of Large Language
  Models}. In \bibinfo{booktitle}{\emph{International Conference on Learning
  Representations (ICLR)}}.
\newblock


\bibitem[Huang et~al\mbox{.}(2017)]%
        {huang2017densenet}
\bibfield{author}{\bibinfo{person}{Gao Huang}, \bibinfo{person}{Zhuang Liu},
  \bibinfo{person}{Laurens van~der Maaten}, {and} \bibinfo{person}{Kilian~Q.
  Weinberger}.} \bibinfo{year}{2017}\natexlab{}.
\newblock \showarticletitle{Densely Connected Convolutional Networks}. In
  \bibinfo{booktitle}{\emph{Proceedings of the IEEE Conference on Computer
  Vision and Pattern Recognition (CVPR)}}. \bibinfo{pages}{4700--4708}.
\newblock
\href{https://doi.org/10.1109/CVPR.2017.243}{doi:\nolinkurl{10.1109/CVPR.2017.243}}


\bibitem[John and Srivastava(1999)]%
        {john1999big}
\bibfield{author}{\bibinfo{person}{Oliver~P. John} {and}
  \bibinfo{person}{Sanjay Srivastava}.} \bibinfo{year}{1999}\natexlab{}.
\newblock \showarticletitle{The Big Five Trait Taxonomy: History, Measurement,
  and Theoretical Perspectives}.
\newblock In \bibinfo{booktitle}{\emph{Handbook of Personality: Theory and
  Research} (\bibinfo{edition}{2nd} ed.)},
  \bibfield{editor}{\bibinfo{person}{Lawrence~A. Pervin} {and}
  \bibinfo{person}{Oliver~P. John}} (Eds.). \bibinfo{publisher}{Guilford
  Press}, \bibinfo{address}{New York}, \bibinfo{pages}{102--138}.
\newblock


\bibitem[Kotov et~al\mbox{.}(2010)]%
        {kotov2010}
\bibfield{author}{\bibinfo{person}{Roman Kotov}, \bibinfo{person}{Wakiza
  Gamez}, \bibinfo{person}{Frank Schmidt}, {and} \bibinfo{person}{David
  Watson}.} \bibinfo{year}{2010}\natexlab{}.
\newblock \showarticletitle{Linking ``Big'' Personality Traits to Anxiety,
  Depressive, and Substance Use Disorders: A Meta-Analysis}.
\newblock \bibinfo{journal}{\emph{Psychological Bulletin}}
  \bibinfo{volume}{136}, \bibinfo{number}{5} (\bibinfo{year}{2010}),
  \bibinfo{pages}{768--821}.
\newblock
\href{https://doi.org/10.1037/a0020327}{doi:\nolinkurl{10.1037/a0020327}}


\bibitem[Kroenke et~al\mbox{.}(2001)]%
        {kroenke2001}
\bibfield{author}{\bibinfo{person}{Kurt Kroenke}, \bibinfo{person}{Robert~L.
  Spitzer}, {and} \bibinfo{person}{Janet B.~W. Williams}.}
  \bibinfo{year}{2001}\natexlab{}.
\newblock \showarticletitle{The {PHQ-9}: Validity of a Brief Depression
  Severity Measure}.
\newblock \bibinfo{journal}{\emph{Journal of General Internal Medicine}}
  \bibinfo{volume}{16}, \bibinfo{number}{9} (\bibinfo{year}{2001}),
  \bibinfo{pages}{606--613}.
\newblock
\href{https://doi.org/10.1046/j.1525-1497.2001.016009606.x}{doi:\nolinkurl{10.1046/j.1525-1497.2001.016009606.x}}


\bibitem[Lin et~al\mbox{.}(2020)]%
        {lin2020bilstm}
\bibfield{author}{\bibinfo{person}{Liyuan Lin}, \bibinfo{person}{Xiaolong
  Chen}, \bibinfo{person}{Ya Shen}, {and} \bibinfo{person}{Lei Zhang}.}
  \bibinfo{year}{2020}\natexlab{}.
\newblock \showarticletitle{Towards Automatic Depression Detection: A
  {BiLSTM/1D CNN}-Based Model}.
\newblock \bibinfo{journal}{\emph{Applied Sciences}} \bibinfo{volume}{10},
  \bibinfo{number}{23} (\bibinfo{year}{2020}), \bibinfo{pages}{8701}.
\newblock
\href{https://doi.org/10.3390/app10238701}{doi:\nolinkurl{10.3390/app10238701}}


\bibitem[Lin(1989)]%
        {lin1989ccc}
\bibfield{author}{\bibinfo{person}{Lawrence I-Kuei Lin}.}
  \bibinfo{year}{1989}\natexlab{}.
\newblock \showarticletitle{A Concordance Correlation Coefficient to Evaluate
  Reproducibility}.
\newblock \bibinfo{journal}{\emph{Biometrics}} \bibinfo{volume}{45},
  \bibinfo{number}{1} (\bibinfo{year}{1989}), \bibinfo{pages}{255--268}.
\newblock
\href{https://doi.org/10.2307/2532051}{doi:\nolinkurl{10.2307/2532051}}


\bibitem[Lin et~al\mbox{.}(2017)]%
        {lin2017}
\bibfield{author}{\bibinfo{person}{Tsung-Yi Lin}, \bibinfo{person}{Priya
  Goyal}, \bibinfo{person}{Ross Girshick}, \bibinfo{person}{Kaiming He}, {and}
  \bibinfo{person}{Piotr Doll\'{a}r}.} \bibinfo{year}{2017}\natexlab{}.
\newblock \showarticletitle{Focal Loss for Dense Object Detection}. In
  \bibinfo{booktitle}{\emph{Proceedings of the IEEE International Conference on
  Computer Vision (ICCV)}}. \bibinfo{pages}{2980--2988}.
\newblock
\href{https://doi.org/10.1109/ICCV.2017.324}{doi:\nolinkurl{10.1109/ICCV.2017.324}}


\bibitem[Liu et~al\mbox{.}(2019)]%
        {liu2019roberta}
\bibfield{author}{\bibinfo{person}{Yinhan Liu}, \bibinfo{person}{Myle Ott},
  \bibinfo{person}{Naman Goyal}, \bibinfo{person}{Jingfei Du},
  \bibinfo{person}{Mandar Joshi}, \bibinfo{person}{Danqi Chen},
  \bibinfo{person}{Omer Levy}, \bibinfo{person}{Mike Lewis},
  \bibinfo{person}{Luke Zettlemoyer}, {and} \bibinfo{person}{Veselin
  Stoyanov}.} \bibinfo{year}{2019}\natexlab{}.
\newblock \showarticletitle{{RoBERTa}: A Robustly Optimised {BERT} Pretraining
  Approach}.
\newblock  (\bibinfo{year}{2019}).
\newblock
\showeprint[arxiv]{1907.11692}~[cs.CL]


\bibitem[Mart{\'i}nez-Nicol{\'a}s et~al\mbox{.}(2025)]%
        {martinez2025elderly}
\bibfield{author}{\bibinfo{person}{Israel Mart{\'i}nez-Nicol{\'a}s},
  \bibinfo{person}{Diego Criado}, \bibinfo{person}{Fernando Gordillo},
  \bibinfo{person}{Francisco Mart{\'i}nez-S{\'a}nchez}, {and}
  \bibinfo{person}{Juan Jos{\'e}~G. Meil{\'a}n}.}
  \bibinfo{year}{2025}\natexlab{}.
\newblock \showarticletitle{Speech Analysis for Detecting Depression in Older
  Adults: A Systematic Review}.
\newblock \bibinfo{journal}{\emph{Frontiers in Psychology}}
  \bibinfo{volume}{16} (\bibinfo{year}{2025}), \bibinfo{pages}{1715538}.
\newblock
\href{https://doi.org/10.3389/fpsyg.2025.1715538}{doi:\nolinkurl{10.3389/fpsyg.2025.1715538}}


\bibitem[Matsushima et~al\mbox{.}(2024)]%
        {matsushima2024development}
\bibfield{author}{\bibinfo{person}{Toshio Matsushima},
  \bibinfo{person}{Yuichiro Yoshikawa}, \bibinfo{person}{Keitaro Matsuo},
  \bibinfo{person}{Keita Kurahara}, \bibinfo{person}{Youki Uehara},
  \bibinfo{person}{Tomohiro Nakao}, \bibinfo{person}{Hiroshi Ishiguro},
  \bibinfo{person}{Hirokazu Kumazaki}, {and} \bibinfo{person}{Takahiro~A.
  Kato}.} \bibinfo{year}{2024}\natexlab{}.
\newblock \showarticletitle{Development of Depression Assessment Tools Using
  Humanoid Robots---Can Tele-Operated Robots Talk with Depressive Persons Like
  Humans?}
\newblock \bibinfo{journal}{\emph{Journal of Psychiatric Research}}
  \bibinfo{volume}{170} (\bibinfo{year}{2024}), \bibinfo{pages}{187--194}.
\newblock
\href{https://doi.org/10.1016/j.jpsychires.2024.01.003}{doi:\nolinkurl{10.1016/j.jpsychires.2024.01.003}}


\bibitem[Patapati et~al\mbox{.}(2025)]%
        {patapati2025daicwoz}
\bibfield{author}{\bibinfo{person}{Santosh~Varma Patapati},
  \bibinfo{person}{Ishan Pendyala}, \bibinfo{person}{Murari Ambati},
  \bibinfo{person}{Pranav Kunadharaju}, \bibinfo{person}{Pranav Kokati},
  \bibinfo{person}{Amit Adiraju}, {and} \bibinfo{person}{Trisanth Srinivasan}.}
  \bibinfo{year}{2025}\natexlab{}.
\newblock \showarticletitle{Most {DAIC-WOZ} Depression Classifiers Are Invalid,
  They Don't Learn Task-Specific Features: Preliminary Findings From a
  Large-Scale Reproducibility Study}. In \bibinfo{booktitle}{\emph{Companion
  Proceedings of the 27th International Conference on Multimodal Interaction
  (ICMI Companion)}}.
\newblock


\bibitem[Radford et~al\mbox{.}(2023)]%
        {radford2023whisper}
\bibfield{author}{\bibinfo{person}{Alec Radford}, \bibinfo{person}{Jong~Wook
  Kim}, \bibinfo{person}{Tao Xu}, \bibinfo{person}{Greg Brockman},
  \bibinfo{person}{Christine McLeavey}, {and} \bibinfo{person}{Ilya
  Sutskever}.} \bibinfo{year}{2023}\natexlab{}.
\newblock \showarticletitle{Robust Speech Recognition via Large-Scale Weak
  Supervision}. In \bibinfo{booktitle}{\emph{Proceedings of the 40th
  International Conference on Machine Learning (ICML)}}.
  \bibinfo{pages}{28492--28518}.
\newblock


\bibitem[Ringeval et~al\mbox{.}(2019)]%
        {ringeval2019}
\bibfield{author}{\bibinfo{person}{Fabien Ringeval}, \bibinfo{person}{Bj\"{o}rn
  Schuller}, \bibinfo{person}{Michel Valstar}, \bibinfo{person}{Nicholas
  Cummins}, \bibinfo{person}{Roddy Cowie}, \bibinfo{person}{Leili Tavabi},
  \bibinfo{person}{Maximilian Schmitt}, \bibinfo{person}{Sina Alisamir},
  \bibinfo{person}{Shahin Amiriparian}, \bibinfo{person}{Eva-Maria Messner},
  \bibinfo{person}{Siyang Song}, \bibinfo{person}{Shuo Liu},
  \bibinfo{person}{Ziping Zhao}, \bibinfo{person}{Adria Mallol-Ragolta},
  \bibinfo{person}{Zhao Ren}, \bibinfo{person}{Mohammad Soleymani}, {and}
  \bibinfo{person}{Maja Pantic}.} \bibinfo{year}{2019}\natexlab{}.
\newblock \showarticletitle{{AVEC} 2019 Workshop and Challenge: State-of-Mind,
  Detecting Depression with AI, and Cross-Cultural Affect Recognition}. In
  \bibinfo{booktitle}{\emph{Proceedings of the 9th International on
  Audio/Visual Emotion Challenge and Workshop}}. \bibinfo{pages}{3--12}.
\newblock
\href{https://doi.org/10.1145/3347320.3357688}{doi:\nolinkurl{10.1145/3347320.3357688}}


\bibitem[Song et~al\mbox{.}(2025)]%
        {song2025survey}
\bibfield{author}{\bibinfo{person}{Siyang Song}, \bibinfo{person}{Yupeng Huo},
  \bibinfo{person}{Shiqing Tang}, \bibinfo{person}{Jiaee Cheong},
  \bibinfo{person}{Rui Gao}, \bibinfo{person}{Michel Valstar}, {and}
  \bibinfo{person}{Hatice Gunes}.} \bibinfo{year}{2025}\natexlab{}.
\newblock \bibinfo{title}{Automatic Depression Assessment using Machine
  Learning: A Comprehensive Survey}.
\newblock
\showeprint[arxiv]{2506.18915}~[cs.LG]


\bibitem[Valstar et~al\mbox{.}(2016)]%
        {avec2016}
\bibfield{author}{\bibinfo{person}{Michel Valstar}, \bibinfo{person}{Jonathan
  Gratch}, \bibinfo{person}{Björn Schuller}, \bibinfo{person}{Fabien
  Ringeval}, \bibinfo{person}{Denis Lalanne}, \bibinfo{person}{Mercedes
  Torres~Torres}, \bibinfo{person}{Stefan Scherer}, \bibinfo{person}{Giota
  Stratou}, \bibinfo{person}{Roddy Cowie}, {and} \bibinfo{person}{Maja
  Pantic}.} \bibinfo{year}{2016}\natexlab{}.
\newblock \showarticletitle{{AVEC} 2016: Depression, Mood, and Emotion
  Recognition Workshop and Challenge}. In \bibinfo{booktitle}{\emph{Proceedings
  of the 6th International Workshop on Audio/Visual Emotion Challenge (AVEC)}}.
  \bibinfo{pages}{3--10}.
\newblock
\href{https://doi.org/10.1145/2988257.2988258}{doi:\nolinkurl{10.1145/2988257.2988258}}


\bibitem[Wang et~al\mbox{.}(2025)]%
        {fosafer}
\bibfield{author}{\bibinfo{person}{Honghong Wang}, \bibinfo{person}{Jing Deng},
  {and} \bibinfo{person}{Rong Zheng}.} \bibinfo{year}{2025}\natexlab{}.
\newblock \bibinfo{title}{Personality-Enhanced Multimodal Depression Detection
  in the Elderly}.
\newblock
\showeprint[arxiv]{2510.08004}~[cs.LG]


\bibitem[Wang et~al\mbox{.}(2023)]%
        {wang2023e5}
\bibfield{author}{\bibinfo{person}{Liang Wang}, \bibinfo{person}{Nan Yang},
  \bibinfo{person}{Xiaolong Huang}, \bibinfo{person}{Linjun Yang},
  \bibinfo{person}{Rangan Majumder}, {and} \bibinfo{person}{Furu Wei}.}
  \bibinfo{year}{2023}\natexlab{}.
\newblock \bibinfo{title}{Improving Text Embeddings with Large Language
  Models}.
\newblock
\showeprint[arxiv]{2401.00368}~[cs.CL]


\bibitem[Wang et~al\mbox{.}(2021)]%
        {wang2021gait}
\bibfield{author}{\bibinfo{person}{Yameng Wang}, \bibinfo{person}{Jingying
  Wang}, \bibinfo{person}{Xiaoqian Liu}, {and} \bibinfo{person}{Tingshao Zhu}.}
  \bibinfo{year}{2021}\natexlab{}.
\newblock \showarticletitle{Detecting Depression Through Gait Data: Examining
  the Contribution of Gait Features in Recognizing Depression}.
\newblock \bibinfo{journal}{\emph{Frontiers in Psychiatry}}
  \bibinfo{volume}{12} (\bibinfo{year}{2021}), \bibinfo{pages}{661213}.
\newblock
\href{https://doi.org/10.3389/fpsyt.2021.661213}{doi:\nolinkurl{10.3389/fpsyt.2021.661213}}


\bibitem[Williamson et~al\mbox{.}(2016)]%
        {williamson2016b}
\bibfield{author}{\bibinfo{person}{James~R. Williamson},
  \bibinfo{person}{Elizabeth Godoy}, \bibinfo{person}{Miriam Cha},
  \bibinfo{person}{Adrianne Schwarzentruber}, \bibinfo{person}{Pooya Khorrami},
  \bibinfo{person}{Youngjune Gwon}, \bibinfo{person}{H.~T. Kung},
  \bibinfo{person}{Charlie~K. Dagli}, {and} \bibinfo{person}{Thomas~F.
  Quatieri}.} \bibinfo{year}{2016}\natexlab{}.
\newblock \showarticletitle{Detecting Depression using Vocal, Facial and
  Semantic Communication Cues}. In \bibinfo{booktitle}{\emph{Proceedings of the
  6th International Workshop on Audio/Visual Emotion Challenge (AVEC)}}.
  \bibinfo{pages}{11--18}.
\newblock
\href{https://doi.org/10.1145/2988257.2988263}{doi:\nolinkurl{10.1145/2988257.2988263}}


\bibitem[{World Health Organization}(2023)]%
        {who2023}
\bibfield{author}{\bibinfo{person}{{World Health Organization}}.}
  \bibinfo{year}{2023}\natexlab{}.
\newblock \bibinfo{title}{Depression Fact Sheet}.
\newblock
  \bibinfo{howpublished}{\url{https://www.who.int/news-room/fact-sheets/detail/depression}}.
\newblock
\shownote{Accessed June 2026}.
\newblock


\bibitem[Xiao et~al\mbox{.}(2025)]%
        {xiao2025restoring}
\bibfield{author}{\bibinfo{person}{Jiancong Xiao}, \bibinfo{person}{Bojian
  Hou}, \bibinfo{person}{Zhanliang Wang}, \bibinfo{person}{Ruochen Jin},
  \bibinfo{person}{Qi Long}, \bibinfo{person}{Weijie Su}, {and}
  \bibinfo{person}{Li Shen}.} \bibinfo{year}{2025}\natexlab{}.
\newblock \showarticletitle{Restoring Calibration for Aligned Large Language
  Models: A Calibration-Aware Fine-Tuning Approach}. In
  \bibinfo{booktitle}{\emph{Proceedings of the 42nd International Conference on
  Machine Learning (ICML)}}.
\newblock


\bibitem[Yang et~al\mbox{.}(2024)]%
        {qwen2024}
\bibfield{author}{\bibinfo{person}{An Yang}, \bibinfo{person}{Baosong Yang},
  \bibinfo{person}{Beichen Zhang}, \bibinfo{person}{Binyuan Hui},
  \bibinfo{person}{Bo Zheng}, \bibinfo{person}{Bowen Yu},
  \bibinfo{person}{Chengyuan Li}, \bibinfo{person}{Dayiheng Liu},
  \bibinfo{person}{Fei Huang}, \bibinfo{person}{Haoran Wei},
  \bibinfo{person}{Huan Lin}, {et~al\mbox{.}}} \bibinfo{year}{2024}\natexlab{}.
\newblock \bibinfo{title}{{Qwen2.5} Technical Report}.
\newblock
\showeprint[arxiv]{2412.15115}~[cs.CL]


\bibitem[Yang et~al\mbox{.}(2022)]%
        {gatortron}
\bibfield{author}{\bibinfo{person}{Xi Yang}, \bibinfo{person}{Aokun Chen},
  \bibinfo{person}{Nima PourNejatian}, \bibinfo{person}{Hoo~Chang Shin},
  \bibinfo{person}{Kaleb~E. Smith}, \bibinfo{person}{Christopher Parisien},
  \bibinfo{person}{Colin Compas}, \bibinfo{person}{Cheryl Martin},
  \bibinfo{person}{Anthony~B. Costa}, \bibinfo{person}{Mona~G. Flores},
  \bibinfo{person}{Ying Zhang}, \bibinfo{person}{Tanja Magoc},
  \bibinfo{person}{Christopher~A. Harle}, \bibinfo{person}{Gloria Lipori},
  \bibinfo{person}{Duane~A. Mitchell}, \bibinfo{person}{William~R. Hogan},
  \bibinfo{person}{Elizabeth~A. Shenkman}, \bibinfo{person}{Jiang Bian}, {and}
  \bibinfo{person}{Yonghui Wu}.} \bibinfo{year}{2022}\natexlab{}.
\newblock \showarticletitle{A Large Language Model for Electronic Health
  Records}.
\newblock \bibinfo{journal}{\emph{npj Digital Medicine}}  \bibinfo{volume}{5}
  (\bibinfo{year}{2022}), \bibinfo{pages}{194}.
\newblock
\href{https://doi.org/10.1038/s41746-022-00742-2}{doi:\nolinkurl{10.1038/s41746-022-00742-2}}


\bibitem[Yates et~al\mbox{.}(2017)]%
        {depresym}
\bibfield{author}{\bibinfo{person}{Andrew Yates}, \bibinfo{person}{Arman
  Cohan}, {and} \bibinfo{person}{Nazli Goharian}.}
  \bibinfo{year}{2017}\natexlab{}.
\newblock \showarticletitle{Depression and Self-Harm Risk Assessment in Online
  Forums}. In \bibinfo{booktitle}{\emph{Proceedings of the 2017 Conference on
  Empirical Methods in Natural Language Processing (EMNLP)}}.
  \bibinfo{pages}{2968--2978}.
\newblock
\href{https://doi.org/10.18653/v1/D17-1322}{doi:\nolinkurl{10.18653/v1/D17-1322}}


\bibitem[Zhang et~al\mbox{.}(2018)]%
        {zhang2018mixup}
\bibfield{author}{\bibinfo{person}{Hongyi Zhang}, \bibinfo{person}{Moustapha
  Cisse}, \bibinfo{person}{Yann~N. Dauphin}, {and} \bibinfo{person}{David
  Lopez-Paz}.} \bibinfo{year}{2018}\natexlab{}.
\newblock \showarticletitle{mixup: Beyond Empirical Risk Minimization}. In
  \bibinfo{booktitle}{\emph{International Conference on Learning
  Representations (ICLR)}}.
\newblock


\bibitem[Zhu et~al\mbox{.}(2023)]%
        {zhu2023calibration}
\bibfield{author}{\bibinfo{person}{Chiwei Zhu}, \bibinfo{person}{Benfeng Xu},
  \bibinfo{person}{Quan Wang}, \bibinfo{person}{Yongdong Zhang}, {and}
  \bibinfo{person}{Zhendong Mao}.} \bibinfo{year}{2023}\natexlab{}.
\newblock \showarticletitle{On the Calibration of Large Language Models and
  Alignment}. In \bibinfo{booktitle}{\emph{Findings of the Association for
  Computational Linguistics: EMNLP 2023}}.
\newblock


\end{thebibliography}

\end{document}